\def\year{2019}\relax
\documentclass[letterpaper]{article} %DO NOT CHANGE THIS
\usepackage{aaai19}  %Required
\usepackage{times}  %Required
\usepackage{helvet}  %Required
\usepackage{courier}  %Required
\usepackage{url}  %Required
\usepackage{graphicx}  %Required

\usepackage{amsmath}
\usepackage{amssymb}
\usepackage{subcaption}
\usepackage{multirow}
\newcommand{\specialcell}[2][c]{%
  \begin{tabular}[#1]{@{}c@{}}#2\end{tabular}}

\begin{document}
\title{Learning to Adaptively Scale Recurrent Neural Networks}
 \author{
 Hao Hu$^1$, Liqiang Wang$^1$ \and  Guo-Jun Qi$^2$ \\
 $^1$University of Central Florida\\
 $^2$Huawei Cloud\\
 \texttt{\{haohu, lwang\}@cs.ucf.edu}, \texttt{Guojun.qi@huawei.com}\\
 }
\maketitle
\begin{abstract}
Recent advancements in recurrent neural network (RNN) research have demonstrated the superiority of utilizing multiscale structures in learning temporal representations of time series. Currently, most of multiscale RNNs use fixed scales, which do not comply with the nature of dynamical temporal patterns among sequences. In this paper, we propose Adaptively Scaled Recurrent Neural Networks (ASRNN), a simple but efficient way to handle this problem. Instead of using predefined scales, ASRNNs are able to learn and adjust scales based on different temporal contexts, making them more flexible in modeling multiscale patterns. Compared with other multiscale RNNs, ASRNNs are bestowed upon dynamical scaling capabilities with much simpler structures, and are easy to be integrated with various RNN cells. The experiments on multiple sequence modeling tasks indicate ASRNNs can efficiently adapt scales based on different sequence contexts and yield better performances than baselines without dynamical scaling abilities.
\end{abstract}

\section{Introduction}\label{sec:introduction}

Recurrent Neural Networks (RNNs) play a critical role in sequential modeling as they have achieved impressive performances in various tasks  \cite{SkipRNN}\cite{DilatedRNN}\cite{HMRNN}\cite{PhasedLSTM}. Yet learning long-term dependencies from long sequences still remains a very difficult task \cite{gradient_vanishing} \cite{gradient_vanishing1}\cite{ye2017temporal}\cite{hu2017temporal}. Among various ways that try to handle this problem, modeling multiscale patterns seem to be a promising strategy since many multiscale RNN structures perform better than other non-scale modeling RNNs in multiple applications \cite{CWRNN}\cite{PhasedLSTM}\cite{HMRNN}\cite{DilatedRNN}\cite{SkipRNN}\cite{chang2014factorized}. Multiscale RNNs can be roughly divided into two groups based on their design philosophies. The first group trends to modeling scale patterns with the hierarchical architectures and prefixed scales for different layers. This may lead to at least two disadvantages. First, the prefixed scale can not be adjusted to fit the temporal dynamics throughout the time. Although patterns in different scale levels require distinct frequencies to update, they do not always stick to a certain scale and could vary at different time steps. For example, in polyphonic music modeling, distinguishing different music styles demands RNNs to model various emotion changes throughout music pieces. While emotion changes are usually controlled by the lasting time of notes, it is insufficient to model such patterns using only fixed scales as the notes last differently at different time. Secondly, stacking multiple RNN layers greatly increases the complexity of the entire model, which makes RNNs even harder to train. Unlike this, another group of multiscale RNNs models scale patterns through gate structures \cite{PhasedLSTM}\cite{SkipRNN}\cite{qi2016hierarchically}. In such cases, additional control gates are learned to optionally update hidden for each time step, resulting in a more flexible sequential representations. Yet such modeling strategy may not remember information which is more important for future outputs but less related to current states.

In this paper, we aim to model the underlying multiscale temporal patterns for time sequences while avoiding all the weaknesses mentioned above. To do so, we present Adaptively Scaled Recurrent Neural Networks (ASRNNs), a simple extension for existing RNN structures, which allows them to adaptively adjust the scale based on temporal contexts at different time steps. Using the causal convolution proposed by \cite{wavenet}, ASRNNs model scale patterns by firstly convolving input sequences with wavelet kernels, resulting in scale-related inputs that parameterized by the scale coefficients from kernels. After that, scale coefficients are sampled from categorical distributions determined by different temporal contexts. This is achieved by sampling Gumbel-Softmax (GM) distributions instead, which are able to approximate true categorical distributions through the re-parameterization trick. Due to the differentiable nature of GM, ASRNNs could learn to flexibly determine which scale is most important to target outputs according to temporal contents at each time step. Compared with other multiscale architectures, the proposed ASRNNs have several advantages. First, there is no fixed scale in the model. The subroutine for scale sampling can be trained to select proper scales to dynamically model the temporal scale patterns. Second, ASRNNs can model multiscale patterns within a single RNN layer, resulting in a much simpler structure and easier optimization process. Besides, ASRNNs do not use gates to control the updates of hidden states. Thus there is no risk of missing information for future outputs. %important information for future outputs.

To verify the effectiveness of ASRNNs, we conduct extensive experiments on various sequence modeling tasks, including low density signal identification, long-term memorization, pixel-to-pixel image classification, music genre recognition and language modeling. Our results suggest that ASRNNs can achieve better performances than their non-adaptively scaled counterparts and are able to adjust scales according to various temporal contents. We organize the rest paper like this: the first following section reviews relative literatures, then we introduce ASRNNs with details in next section; after that the results for all evaluations are presented, and the last section concludes the paper.

\section{Related Work}\label{sec:related}

As a long-lasting research topic, the difficulties of training RNNs to learn long-term dependencies are considered to be caused by several reasons. First, the gradient exploding and vanishing problems during back propagation make training RNNs very tough \cite{gradient_vanishing} \cite{gradient_vanishing1}. Secondly, RNN memory cells usually need to keep both long-term dependencies and short-term memories simultaneously, which means there should always be trade-offs between two types of information.  To overcome such problems, some efforts aim to design more sophisticated memory cell structures. For example, Long-short term memory (LSTM) \cite{LSTM} and gated recurrent unit (GRU) \cite{GRU}, are able to capture more temporal information; while some others attempt to develop better training algorithms and initialization strategies such as gradient clipping \cite{gradient_clip}, orthogonal and unitary weight optimization \cite{uRNN}\cite{init_rlu} \cite{full_uRNN}\cite{qi2009learning}\cite{wang2016supervised}\cite{qi2012clustering} etc. These techniques can alleviate the problem to some extent \cite{tang2017tri}\cite{li2017linear}\cite{wang2012recommending}.

Meanwhile, previous works like \cite{CWRNN} \cite{PhasedLSTM} \cite{HMRNN} \cite{tang2007typicality} \cite{hua2008online} suggest learning temporal scale structures is also the key to this problem. This stands upon the fact that temporal data usually contains rich underlying multiscale patterns \cite{schmidhuber1991neural}\cite{mozer1992induction} \cite{HRNN} \cite{lin1996learning} \cite{SFM}. To model multiscale patterns, a popular strategy is to build hierarchical architectures. These RNNs such as hierarchical RNNs \cite{HRNN}, clockwork RNNs \cite{CWRNN} and Dilated RNNs \cite{DilatedRNN} etc, contain hierarchical architectures whose neurons in high-level layers are less frequently updated than those in low-level layers. Such properties fit the natures of many latent multiscale temporal patterns where low-level patterns are sensitive to local changes while high-level patterns are more coherent with the temporal consistencies. Instead of considering hierarchical architectures, some multiscale RNNs model scale patterns using control gates to decide whether to update hidden states or not at a certain time step. Such structures like phased LSTMs \cite{PhasedLSTM} and skip RNNs \cite{SkipRNN}, are able to adjust their modeling scales based on current temporal contexts, leading to more reasonable and flexible sequential representations. Recently, some multiscale RNNs like hierarchical multi-scale RNNs \cite{HMRNN}, manage to combine the gate-controlling updating mechanism into hierarchical architectures and has made impressive progress in language modeling tasks. Yet they still employ multi-layer structures which make the optimization not be easy.

\section{Adaptively Scaled Recurrent Neural Networks} \label{sec:ASRNNs}

In this section we introduce Adaptively Scaled Recurrent Neural Networks (ASRNNs), a simple but useful extension for various RNN cells that allows to dynamically adjust scales at each time step. An ASRNNs is consist of three components: scale parameterization, adaptive scale learning and RNN cell integration, which will be covered in following subsections.

\subsection{Scale Parameterization}\label{subsec:scale_para}

We begin our introduction for ASRNNs with scale parameterization. Suppose $\mathbf{X} = [\mathbf{x}_1, \mathbf{x}_2 \cdots, \mathbf{x}_T] $ is an input sequence where $\mathbf{x}_t \in \mathcal{R}^{n}$. At time $t$, instead of taking only the current frame $\mathbf{x}_t$ as input, ASRNNs compute an alternative scale-related input $\mathbf{\tilde{x}}_t$, which can be obtained by taking a causal convolution between the original input sequence $\mathbf{X}$ and a scaled wavelet kernel function $\phi_{j_t}$.

More specifically, let $J$ be the number of considered scales. Consider a wavelet kernel $\phi$ of size $K$. At any time $t$, given a scale $j_t \in \{0, \cdots, J-1\}$, the input sequence $\mathbf X$ is convolved with a scaled wavelet kernel $\phi_{j_t}=\phi(\frac{i}{2^{j_t}})$. This yields the following scaled-related input $\mathbf{\tilde{x}}_t$ at time $t$
\begin{equation}\label{eq:tilde_x}
	\mathbf{\tilde{x}}_t = (\mathbf{X} * \phi_{j_t})_t = \sum_{i =0}^{2^{j_t}K-1} \mathbf{x}_{t-i}\phi(\frac{i}{2^{j_t}}) \in \mathcal{R}^n
\end{equation}
where for any $i \in \{t-2^{j_t}K+1, \cdots, t-1\}$, we manually set $\mathbf{x}_{i}= \mathbf{0}$ iff $i \leq 0$. And the causal convolution operator $*$  \cite{wavenet}  is defined to avoid the resultant $\mathbf{\tilde{x}}_t$  depending on future inputs. We also let $\phi(\frac{i}{2^{j_t}})=0$ iff $2^{j_t} \nmid i$. It is easy to see that $\mathbf{\tilde{x}}_t$ can only contain information from $\mathbf{x}_{t-i}$ when $i = 2^{j_t}k, k \in \{1, \cdots, K\}$. In other words, there are skip connections between $\mathbf{x}_{t-2^{j_t}(k-1)}$ and $\mathbf{x}_{t-2^{j_t}k}$ in the scale $j_t$. While $j_t$ becomes larger, the connections skip further.

It is worth mentioning that the progress for obtaining scale-related input $\mathbf{\tilde{x}}_t$ is quite similar as the convolutions with the real waveforms in \cite{wavenet}. By stacking several causal convolutional layers, \cite{wavenet} is able to model temporal patterns in multiple scale levels with its exponential-growing receptive field. However, such abilities are achieved through a hierarchical structure where each layer is given a fixed dilation factor that does not change through out time. To avoid this, we replace the usual convolution kernels with wavelet kernels, which come with scaling coefficients just like $j_t$ in equation \ref{eq:tilde_x}. By varying $j_t$, $\mathbf{\tilde{x}}_t$ is allowed to contain information from different scale levels. Thus we call it scale parameterization. We further demonstrate it's possible to adaptively control $j_t$ based on temporal contexts through learning, which will be discussed in subsection \ref{subsec:scale_learn}.

\subsection{Adaptive Scale Learning}\label{subsec:scale_learn}

To adjust scale $j_t$ at different time $t$, we need to sample $j_t$ from a categorical distribution where each class probability is implicitly determined by temporal contexts. However, it is impossible to directly train such distributions along with deep neural networks because of the non-differentiable nature of their discrete variables. Fortunately, \cite{gumbelsoftmax} \cite{gumbelsoftmax2} propose Gumbel-Softmax (GM) distribution, a differentiable approximation for a categorical distribution that allow gradients to be back propagated through its samples. Moreover, GM employs the re-parameterization trick, which divides the distribution into a basic independent random variable and a deterministic function. Thus, by learning the function only, we can bridge the categorical sampling with temporal contexts through a differentiable process.

Now we introduce the process of learning to sample scale $j_t$ with more details. Suppose $\boldsymbol{\pi}_t = [\pi_0^t, \cdots, \pi_{J-1}^t] \in [0, 1]^J$ are class probabilities for scale set $\{0, \cdots, J-1\}$ and $\mathbf{z}_t = [z_0^t, \cdots, z_{J-1}^t] \in \mathcal{R}^J$ are some logits related to temporal contexts at time $t$. The relationship between $\boldsymbol{\pi}_t$ and $\mathbf{z}_t$ can be written as
\begin{equation}\label{eq:pi_t}
	\pi_i^t = \frac{\exp(z_i^t)}{\sum_{i'=0}^{J-1} \exp(z_{i'}^t)}
\end{equation}
where $i \in \{0, \cdots, J-1\}$. Let $\mathbf{y}_t = [y_0^t, \cdots, y_{J-1}^t] \in [0,1]^J$ be a sample from GM. Based on \cite{gumbelsoftmax}, $y_i^t$ for $i=0, \cdots, J-1$ can be calculated as
\begin{equation}\label{eq:y_t}
	y_i^t = \frac{\exp((\log\pi_i^t+g_i)/\tau)}{\sum_{i'=0}^{J-1} \exp((\log\pi_{i'}^t+g_{i'})/\tau)}
\end{equation}
where $g_0, \cdots, g_{J-1}$ are i.i.d. samples drawn from the basic Gumbel$(0, 1)$ distribution and $\tau$ controls how much the GM is close to a true categorical distribution. In other words, as $\tau$ goes to $0$, $\mathbf{y}_t$ would become $\mathbf{j}_t$, the one-hot vector whose $j_t$th value is $1$.

Thus with GM, it is clear that the sampled $\mathbf{j}_t$ is approximated by a differentiable function of $\mathbf{z}_t$. We further define $\mathbf{z}_t$ with the hidden states $\mathbf{h}_{t-1} \in \mathcal{R}^m$ and input $\mathbf{x}_t \in \mathcal{R}^n$ as
\begin{equation}\label{eq:z_t}
	\mathbf{z}_t = \mathbf{W}_z\mathbf{h}_{t-1}+\mathbf{U}_z \mathbf{x}_t+\mathbf{b}_z \in \mathcal{R}^J
\end{equation}
where $\mathbf{W}_z, \mathbf{U}_z$ are weight matrices and $\mathbf{b}_z$ is bias vector. Combing equations \ref{eq:pi_t}, \ref{eq:y_t} and \ref{eq:z_t}, we achieve our goal of dynamically changing $j_t$ by sampling from GM distributions that parameterized by $\mathbf{h}_{t-1}$ and $\mathbf{x}_t$. Since the entire procedure is differentiable, matrices $\mathbf{W}_z$ and $\mathbf{U}_z$ can be optimized along with other parameters of ASRNNs during the training.

\subsection{Integrating with Different RNN Cells}\label{subsec:integrate}

With both the techniques introduced in previously introduced two subsections, we are ready to incorporate the proposed adaptive scaling mechanism with different RNN cells, resulting in various forms  of ASRNNs. Since both $\mathbf{\tilde{x}}_t$ and sampling for $j_t$ don't rely on any specific memory cell designs, it's straightforward to do so by replacing original input frames $\mathbf{x}_t$ with $\mathbf{\tilde{x}}_t$. For example, a ASRNN with LSTM cells can be represented as
\begin{equation}\label{eq:lstm_gates}
	\mathbf{f}_t, \mathbf{i}_t, \mathbf{o}_t= \mathrm{sigmoid}(\mathbf{W}_{f,i,o}\mathbf{h}_{t-1}+\mathbf{U}_{f,i,o} \mathbf{\tilde{x}}_t+\mathbf{b}_{f,i,o}) \in \mathcal{R}^m
\end{equation}
\begin{equation}\label{eq:lstm_g}
	\mathbf{g}_t = \tanh(\mathbf{W}_{g}\mathbf{h}_{t-1}+\mathbf{U}_{g} \mathbf{\tilde{x}}_t+\mathbf{b}_{g}) \in \mathcal{R}^m
\end{equation}
\begin{equation}\label{eq:lstm_c}
	\mathbf{c}_t = \mathbf{f}_t \circ \mathbf{c}_{t-1}+\mathbf{i}_t \circ \mathbf{g}_t
\end{equation}
\begin{equation}\label{eq:lstm_h}
	\mathbf{h}_t = \mathbf{o}_t \circ \tanh(\mathbf{c}_t)
\end{equation}
while a ASRNN with GRU cells can be written as
\begin{equation}\label{eq:gru_gates}
	\mathbf{z}_t, \mathbf{r}_t = \mathrm{sigmoid}(\mathbf{W}_{z,r}\mathbf{h}_{t-1}+\mathbf{U}_{z,r} \mathbf{\tilde{x}}_t+\mathbf{b}_{z,r}) \in \mathcal{R}^m
\end{equation}
\begin{equation}\label{eq:gru_g}
	\mathbf{g}_t = \tanh(\mathbf{W}_{g}(\mathbf{r}_t \circ \mathbf{h}_{t-1})+\mathbf{U}_{g} \mathbf{\tilde{x}}_t+\mathbf{b}_{g}) \in \mathcal{R}^m
\end{equation}
\begin{equation}\label{eq:gru_h}
	\mathbf{h}_t = \mathbf{z}_t \circ \mathbf{h}_{t-1}+(1-\mathbf{z}_t) \circ \mathbf{g}_t
\end{equation}

where $\mathbf{W}_*, \mathbf{U}_*$ are weight matrices and $\mathbf{b}_*$ are bias vectors, and $\circ$ means element-wise multiplication. For rest of this paper, we use ASLSTMs to refer those integrated with LSTM cells, ASGRUs for those integrated with GRU cells and so on and so forth. We still call them ASRNNs when there is no specified cell types. It is also worth mentioning that a conventional RNN cell is the special case of its ASRNN counterpart when $J=K=1$.

\subsection{Discussion}\label{subsec:discusion}

Finally, we briefly analyze the advantages of ASRNNs over other multiscale RNN structures. As mentioned in section \ref{sec:introduction}, there are many RNNs, including hierarchical RNNs \cite{HRNN} and Dilated RNNs \cite{DilatedRNN} etc, that apply hierarchical architectures to model multiscale patterns. Compared to them, the advantages of ASRNNs are clear. First, ASRNNs are able to model patterns with multiple scale levels within a single layer, making their spatial complexity much lower than hierarchical structures. Although hierarchical models may reduce the neuron numbers for each layer to have the similar size as single layer ASRNNs, they are harder to train with deeper structures. What's more, compared with the fixed scales for different layers, adapted scales are easier to capture underlying patterns as they can be adjusted based on temporal contexts at different time steps.

Besides, other multiscale RNN models like phased LSTMs \cite{PhasedLSTM} and skip RNNs \cite{SkipRNN} etc, build gate structures to manage scales. Such gates are learned to determine whether to remember the incoming information at each time. However, this may lose information which is important for future time but not for current time. This problem would never happen to ASRNNs as according to equation \ref{eq:tilde_x}, the current input $\mathbf{x}_t$ will always be included in $\mathbf{\tilde{x}}$ and $\mathbf{h}_t$ is updated every step. Thus there is no risk for ASRNNs to lose critical information. This is an important property especially for tasks with frame labels. In such cases previously irrelevant information may become necessary for later frame outputs. Thus information from every frame should be leveraged to get correct outputs at different time.

\section{Experiments}\label{sec:exp}

\begin{table}[t]
\caption{Accuracies for ASRNNs and baselines .}
\label{tab:signal_type_acc}

\begin{center}
\begin{small}
\begin{sc}
\begin{tabular}{lccc}
\hline
Accuracy (\%) & RNN & SRNN & ASRNN \\
\hline\hline
%\abovespace
LSTM & $81.3$ & $83.6$ & $97.7$ \\
GRU & $84.1$ & $88.1$ & $98.0$ \\
\hline

\end{tabular}
\end{sc}
\end{small}
\end{center}
%\vskip -0.2in
\end{table}

\begin{figure*}
    \centering
    \begin{subfigure}[b]{0.49\textwidth}
        \includegraphics[width=\textwidth]{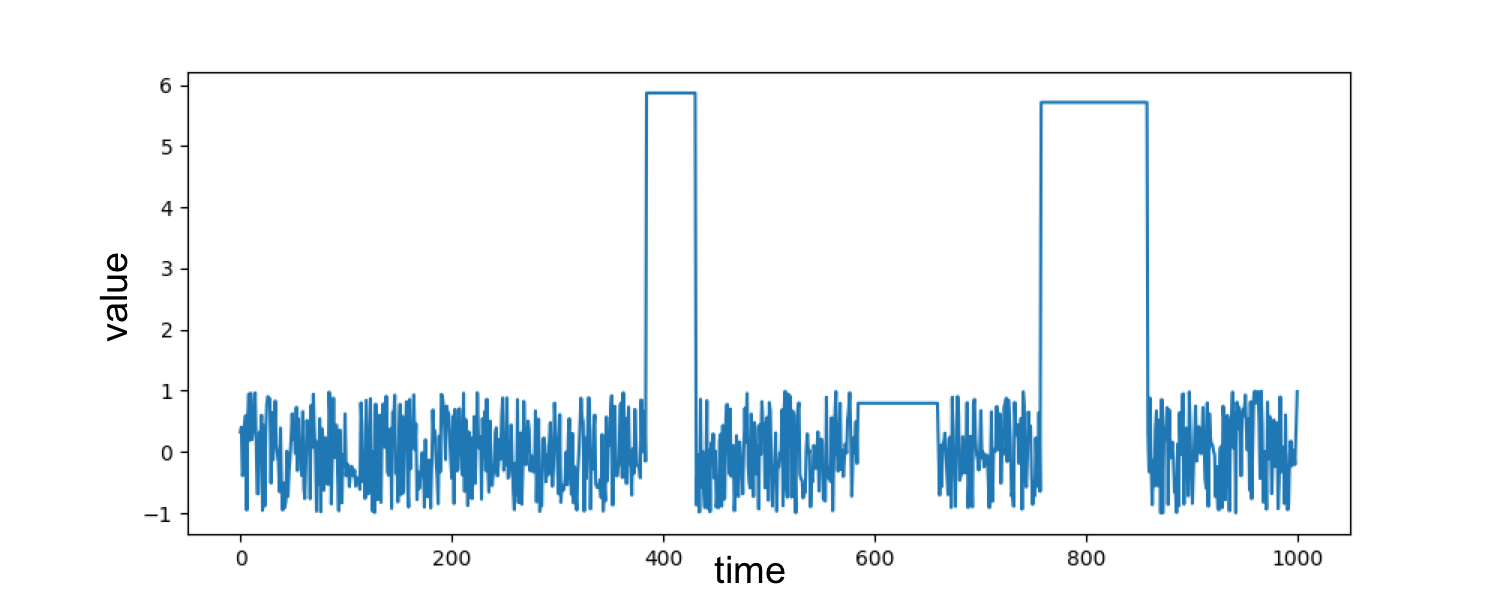}
        %\caption{Accuracies for signal type predictions }
        \caption{A square wave sample.}
        \label{fig:signal_type_acc}
    \end{subfigure}
    \begin{subfigure}[b]{0.49\textwidth}
        \includegraphics[width=\textwidth]{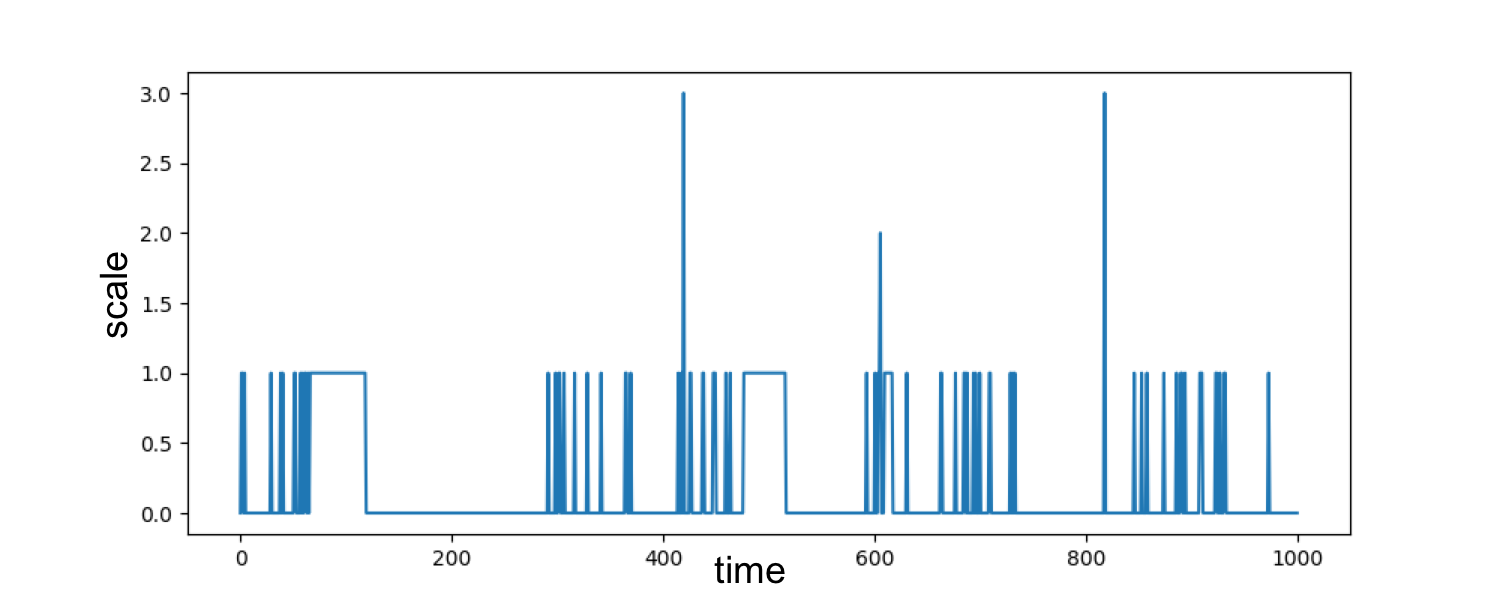}
        %\caption{A square wave sample (top) and corresponding scale variations (bottom)}
        \caption{The corresponding scale variations.}
        \label{fig:signal_type_pattern}
    \end{subfigure}
    %\caption{Results for low density signal type identifications. Figure \ref{fig:signal_type_acc} is the accuracies for ASRNNs and baselines, while figure \ref{fig:signal_type_pattern} shows similarities between a raw square wave and its scale variations.}
    \caption{The similar patterns between a raw square wave and its scale variations.}
    \label{fig:signal_type}
\end{figure*}

In this section, we evaluate the proposed ASRNNs with five sequence modeling tasks: low density signal type identification, copy memory problem, pixel-to-pixel image classification, music genre recognition and word level language modeling. We also explore how the scales would be adapted along time. Unless specified otherwise, all the models are implemented using Tensorflow library \cite{tensorflow}. We train all the models with the RMSProp optimizer \cite{RMSProp} and set learning rate and decay rate to $0.001$ and $0.9$, respectively. It is worth mentioning that there is no techniques such as recurrent batch norm \cite{recurrentbatchnorm} and gradient clipping \cite{gradient_clip} applied during the training. All the weight matrices are initialized with glorot uniform initialization \cite{glorot_init}. For ASRNNs, we choose Haar wavelet as default wavelet kernels, and set $\tau$ of Gumbel-Softmax to $0.1$. We integrate ASRNNs with two popular RNN cells, LSTM \cite{LSTM} and GRU \cite{GRU} and use their conventional counterparts as common baselines. Besides, the baselines also include scaled RNNs (SRNNs), a simplified version that every $j_t$ is set to $J-1$. Additional baselines for individual tasks will be stated in the corresponding subsections if there are. For both SRNNs and ASRNNs, The maximal considered scale $J$ and wavelet kernel size $K$ are set to $4$ and $8$, respectively.

\subsection{Low Density Signal Type Identification}

We begin our evaluation for ASRNNs with some synthetic data. The first task is low density signal type identification, which demands RNNs to distinguish the type of a long sequence that only contains limited useful information. More specifically, consider a sequence with length of $1000$, first we randomly choose $p$ subsequences at arbitrary locations of the sequence where $p \in \{3, 4, 5\}$. Each subsequence has different length $T$ where $T \in \mathbb{Z}^+ \cap [20, 100]$ and we make sure that subsequences don't overlap with each other. For one sequence, all of its subsequences belong to one of the three types of waves: square wave, saw-tooth wave and sine wave, but with different amplitude $A$ sampled from $[-7, 7]$. The rests of the sequence are filled with random noises sampled from $(-1, 1)$. The target is to identify which type of wave a sequence contains. Apparently, a sequence carries only $6\% \sim 50\%$ useful information, requiring RNNs capable of locating it efficiently.

Following above criterion, we randomly generate $2000$ low density sequences for each type. We choose $1600$ sequences per type for training and the remaining are for testing. Table \ref{tab:signal_type_acc} demonstrates the identification accuracies for baselines and ASRNNs. We can see the accuracies of both ASLSTM and ASGRU are over $97.5\%$, meaning they have correctly identified the types for most of sequences without being moderated by noise. Considering the much lower performance of baselines, it's confident to say that ASRNNs are able to efficiently locate useful information with adapted scales. Besides, we also observe there are similar patterns among some waves and their scale variation sequences. Figure \ref{fig:signal_type} gives such an example, from which we see the scale 0 and 1 are more related to noises while the scale 2 and 3 only appear in the region with square form information. Moreover, the subsequence where the scale 2 is located is harder to identify as its values are too close to the noise. We believe such phenomena implies the scale variations could reflect some certain aspects that are helpful for understanding underlying temporal patterns.

\subsection{Copy Memory Problem}

\begin{figure}
    \centering
    \begin{subfigure}[b]{0.23\textwidth}
        \includegraphics[width=\textwidth]{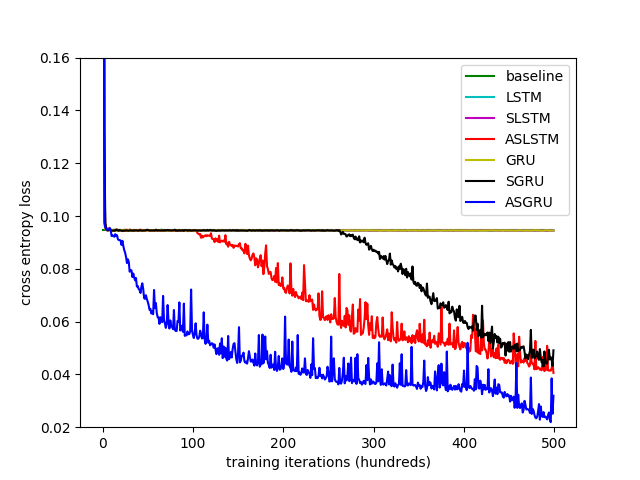}
        \caption{$T = 200$ }
        \label{fig:copy_200}
    \end{subfigure}%\hspace{3mm}
    \begin{subfigure}[b]{0.23\textwidth}
        \includegraphics[width=\textwidth]{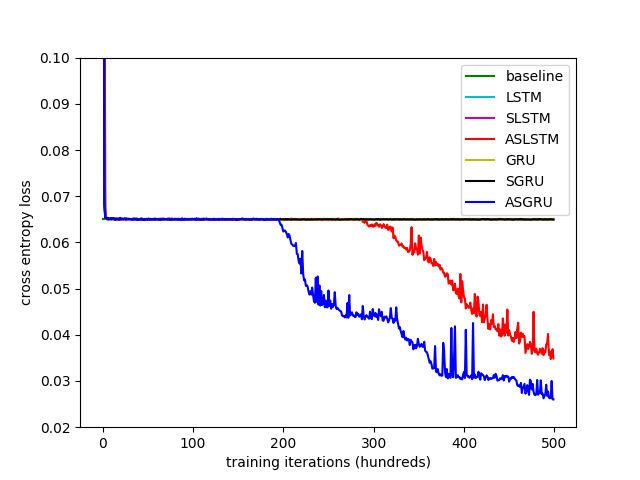}
        \caption{$T = 300$}
        \label{fig:copy_300}
    \end{subfigure}
    \caption{Cross entropies for copy memory problem. Best viewed in colors.}
    \label{fig:copy}
\end{figure}

Next we revisit the copy memory problem, one of the original LSTM tasks proposed by \cite{LSTM} to test the long-term dependency memorization abilities for RNNs. We closely follow the experimental setups used in \cite{uRNN} \cite{full_uRNN}. For each input sequence with $T+20$ elements, The first ten are randomly sampled from integers $0$ to $7$. Then the rest of elements are all set to $8$ except the $T+10$th to $9$, indicating RNNs should begin to replicate the first $10$ elements from now on. The last ten values of output sequence should be exactly the same as the first ten of the input. Cross entropy loss is applied for each time step. In addition to common baselines, we also adopt the memoryless baseline proposed by \cite{uRNN}. The cross entropy of this baseline is $\frac{10\log(8)}{T+20}$, which means it always predict $8$ for first $T+10$ steps while give a random guess of $0$ to $7$ for last 10 steps. For each $T$, we generate $10000$ samples to train all RNN models.

Figure \ref{fig:copy} demonstrates the cross entropy curves of baselines and ASRNNs. We notice that both LSTM and GRU get stuck at the same cross entropy level with the memoryless baseline during the entire training process for both $T=200$ and $T=300$, indicating both LSTM and GRU are incapable of solving the problem with long time delays. This also agrees with the results reported in \cite{NTM} and \cite{uRNN}. For SRNNs, it seems like fixed scales are little helpful since only the SGRU at $T=200$ can have a lower entropy after 250 hundred steps. Unlike them, cross entropies of ASRNNs are observed to further decrease after certain steps of staying with baselines. Especially for $T=200$, ASGRU almost immediately gets the entropy below the baseline with only a few hundreds of iterations passed. Besides, comparing figure \ref{fig:copy_200} and \ref{fig:copy_300}, ASGRUs are more resistant to the increasing of $T$ as ASLSTMs need more time to wait before they can further reduce cross entropies. Overall, such behaviors prove ASRNNs have stronger abilities for memorizing long-term dependencies than baselines.

\subsection{Pixel-to-Pixel Image Classification}\label{subsec:mnist}

\begin{table*}[t]
\caption{Classification accuracies for pixel-to-pixel MNIST. $N$ stands for the number of hidden states. Italic numbers are results reported in the original papers. Bold numbers are best results for each part. ACC=accuracy, UNP/PER=unpermuted/permuted.}
\label{tab:MNIST_acc}
%\vskip 0.1in
\begin{center}
\begin{small}
\begin{sc}
\begin{tabular}{lccccccc}
\hline
%\abovespace\belowspace
RNN & $N$ & \specialcell{\# of\\weights} & \specialcell{Min.\\Scale} & \specialcell{Max.\\Scale} & \specialcell{Avg.\\Scale} & \specialcell{Unp\\Acc(\%)}  & \specialcell{Per\\Acc(\%)}\\
\hline
%\abovespace
LSTM & $129$ & $\approx 68$k & $0$ & $0$ & $0$ & $97.1$ & $89.3$\\
SLSTM & $129$ & $\approx 68$k & $3$ & $3$ & $3$ & $97.4$ &$87.7$\\
ASLSTM & $128$ & $\approx 68$k & $0$ & $3$ & $0.92$ & $\mathbf{98.3}$ & $\mathbf{90.8}$\\
\hline
GRU & $129$ & $\approx 51$k & $0$ & $0$ & $0$ & $96.4$ & $90.1$\\
SGRU & $129$ & $\approx 51$k & $3$ & $3$ & $3$ & $97.0$ & $89.8$\\
ASGRU & $128$ & $\approx 51$k & $0$ & $3$ & $0.75$ & $\mathbf{98.1}$ & $\mathbf{91.2}$\\
\hline
TANH-RNN \cite{init_rlu} & $100$ & - & - & - & - & $\mathit{35.0}$ & $\mathit{33.0}$\\
uRNN \cite{uRNN} & $512$ & $\approx 16$k & - & - & - & $\mathit{95.1}$ & $\mathit{91.4}$\\
\specialcell{Full-capacity \\uRNN \cite{full_uRNN} } & $512$ & $\approx 270$k & - & - & - & $\mathit{96.9}$ & $\mathit{94.1}$\\
iRNN \cite{init_rlu} & $100$ & - & - & - & - & $\mathit{97.0}$ & $\mathit{82.0}$ \\
Skip-LSTM \cite{SkipRNN} & $110$ & - & - & - & - & $\mathit{97.3}$ & -\\
Skip-GRU \cite{SkipRNN} & $110$ & - & - & - & - & $\mathit{97.6}$ & -\\
sTANH-RNN  \cite{sTanhRNN} & $64$ & - & - & - & - & $\mathit{98.1}$ & $\mathit{94.0}$\\
\specialcell{recurrent \\BN-RNN  \cite{RBN}} & $100$ & - & - & - & - & $\mathit{\mathbf{99.0}}$ & $\mathit{\mathbf{95.4}}$\\
\hline
\end{tabular}
\end{sc}
\end{small}
\end{center}
%\vskip -0.1in
\end{table*}

Now we proceed our evaluation for ASRNNs with real world data. In this subsection, we study the pixel-to-pixel image classification problem using MNIST benchmark \cite{MNIST}. Initially proposed by \cite{init_rlu}, it reshapes all $28\times 28$ images into pixel sequences with length of $784$ before fed into RNN models, resulting in a challenge task where capturing long term dependencies is critical. We follow the standard data split settings and only feed outputs from the last hidden state to a linear classifier \cite{xing2010brief}. We conduct experiments for both unpermuted and permuted settings.

%The objective is to minimize the cross entropy between ground truth and predictions.

Table \ref{tab:MNIST_acc} summarizes results of all experiments for pixel-to-pixel MNIST classifications. The first two blocks are the comparisons between common baselines and ASRNNs with different cell structures. Their numbers of weights are adjusted to keep approximately same in order to be compared fairly. We also include other state-of-the-art results of single layer RNNs in the third block. It is easy to see that both SRNNs and ASRNNs achieve better performances than conventional RNNs with scale-related inputs on both settings. This is probably because causal convolutions between inputs and wavelet kernels can be treated as a spatial convolutional layer, allowing SRNNs and ASRNNs to leverage information that is spatially local but temporally remote. Moreover, the adapted scales help ASRNNs further reach the state-of-the-art performances by taking dilated convolutions with those pixels that more spatially related to the current position. It is also worth mentioning the proposed dynamical scaling is totally compatible with the techniques from the third part of the table \ref{tab:MNIST_acc} such as recurrent batch normalization \cite{RBN} and recurrent skip coefficients \cite{sTanhRNN}. Thus ASRNNs can also benefit from them as well.

\subsection{Music Genre Recognition}

\begin{table*}[t]
\caption{Music genre recognition on FMA-small. $N$ stands for the number of hidden states. ACC=accuracy.}
\label{tab:MGR_acc}
%\vskip 0.1in
\begin{center}
\begin{small}
\begin{sc}
\begin{tabular}{llcccccc}
\hline
%\abovespace\belowspace
Features & Methods & $N$ & \specialcell{\# of\\weights} & \specialcell{Min.\\Scale} & \specialcell{Max.\\Scale} & \specialcell{Avg.\\Scale} & Acc(\%)  \\
\hline\hline
%\abovespace
\multirow{7}{*}{MFCC} & LSTM & $129$ & $\approx 74$k & $0$ & $0$ & $0$ & $37.1$ \\
& SLSTM & $129$ & $\approx 74$k & $3$ & $3$ & $3$ & $37.7$ \\
& ASLSTM & $128$ & $\approx 74$k & $0$ & $3$ & $1.34$ & $\mathbf{40.9}$ \\
\cline{2-8}
& GRU & $129$ & $\approx 56$k & $0$ & $0$ & $0$ & $38.2$ \\
& SGRU & $129$ & $\approx 56$k & $3$ & $3$ & $3$ & $38.5$ \\
& ASGRU & $128$ & $\approx 56$k & $0$ & $3$ & $1.39$ & $\mathbf{42.4}$ \\
\cline{2-8}
& MFCC+GMM \cite{music_GMM} & - & - & - & - & - & $21.3$ \\
%& MFCC+SVM \cite{music_GMM} & - & - & - & - & - & $35.6$ \\
\hline\hline
\multirow{7}{*}{Raw} & LSTM & $129$ & $\approx 68$k & $0$ & $0$ & $0$ & $18.5$ \\
& SLSTM & $129$ & $\approx 68$k & $3$ & $3$ & $3$ & $18.9$ \\
& ASLSTM & $128$ & $\approx 68$k & $0$ & $3$ & $1.47$ & $\mathbf{20.1}$ \\
\cline{2-8}
& GRU & $129$ & $\approx 51$k & $0$ & $0$ & $0$ & $18.8$ \\
& SGRU & $129$ & $\approx 51$k & $3$ & $3$ & $3$ & $18.4$ \\
& ASGRU & $128$ & $\approx 51$k & $0$ & $3$ & $1.59$ & $\mathbf{19.5}$ \\
\cline{2-8}
& Raw+CNN \cite{music_CNN} & - & - & - & - & - & $17.5$ \\
\hline

\end{tabular}
\end{sc}
\end{small}
\end{center}
%\vskip -0.1in
\end{table*}

\begin{figure}
  \centering
    \includegraphics[width=.45\textwidth]{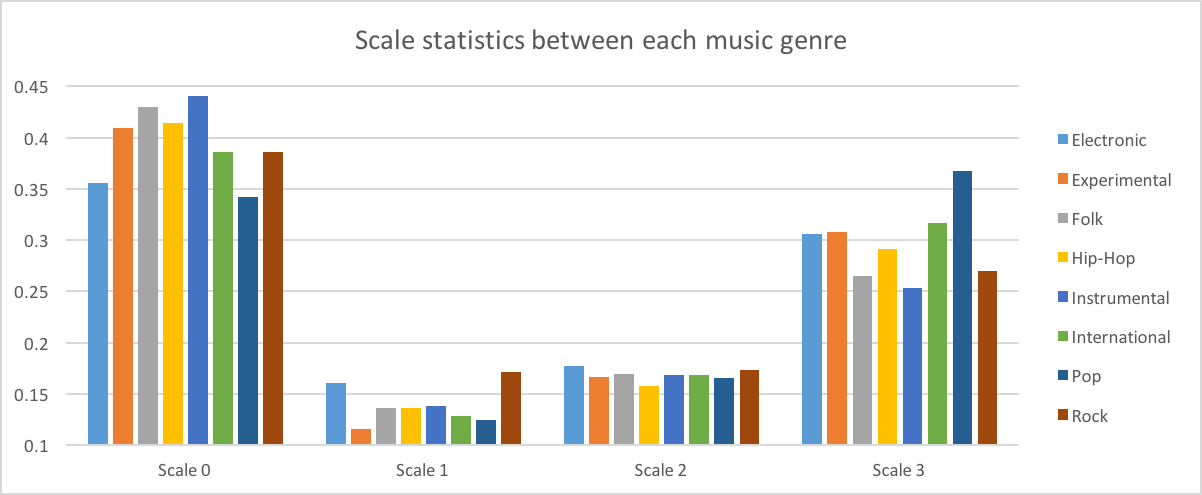}
   \caption{Statistics of scale selections between each music genre. The height of each bar indicates the ratio of how much times the scale is selected in the corresponding genre. Best viewed in colors.}
\label{fig:fma_scale_stat}
\end{figure}

The next evaluation mission for ASRNNs is music genre recognition (MGR), a critical problem in the music information retrieval (MIR) \cite{mckay2006musical} which requires RNNs to characterize the similarities between music tracks across many aspects such as cultures, artists and ages. Compared to other acoustic modeling tasks like speech recognition, MGR is considered to be more difficult as the boundaries between genres are hard to distinguish due to different subjective feelings among people \cite{scaringella2006automatic}. We choose free music archive (FMA) dataset \cite{FMA} to conduct our experiments. More specifically, we use the FMA-small, a balanced FMA subset containing $8000$ music clips that distributed across $8$ genres, where each clip lasts $30$ seconds with sampling rate of $44100$ Hz. We follow the standard $80/10/10\%$ data splitting protocols to get training, validation and test sets. We compute $13$-dimensional log-mel frequency features (MFCC) with $25$ms windows and $10$ms frame steps for each clip, resulting in very long sequences with about $3000$ entries. Besides, inspired by recent success of \cite{wavenet} and \cite{cldnn}, we are also encouraged to directly employ raw audio waves as inputs. Due to limited computational resources, we have to reduce the sampling rate to $200$ Hz for raw music clips while resultant sequences are still two times longer than MFCC sequences.

We demonstrate all the MGR results on FMA-small in the Table \ref{tab:MGR_acc}. Besides RNN models, we also include two baselines without temporal modeling abilities (GMM for MFCC and CNN for raw). We can see when using MFCC features, both the ASLSTM and ASGRU can outperform SRNNs and their conventional counterparts with about $3 \sim 4\%$ improvements. This is an encouraging evidence to show how adapted scales can boost the modeling capabilities of RNNs for MGR. However, the recognition accuracies drop significantly for all models when applying raw audio waves as inputs. In such cases, the gains from adapted scales are marginal for both the ASLSTM and ASGRU. We believe it is due to the low sampling rate for raw music clips since too much information is lost. However, increasing sampling rate will significantly rise the computational costs and make it eventually prohibitive for training RNNs. %Thus, this could be a non-trivial problem and worth further investigations.

%\begin{figure*}
%  \centering
%    \includegraphics[width=\textwidth]{sorted_lstm.png}
%   \caption{Scale variations for every music track in test set of FMA-small using MFCC features. Each genre contains $100$ distinct tracks. Best viewed in colors.}
%\label{fig:fma_scale}
%\end{figure*}

%What's more, figure \ref{fig:fma_scale} visualizes the scale variations across time steps for all testing music tracks using MFCC features. We can see scales are changing rapidly across the entire set.

To further understand the patterns behind such variations, we do statistics on how many times a scale has been selected for each genre, which is normalized and illustrated in figure \ref{fig:fma_scale_stat}. In general, all genres prefer to choose scale 0 and 3 since their ratio values are significantly higher than the other two. However, there are also obvious differences between genres within the same scale. For example, instrumental music tracks have more steps with scale 0 than Pop musics, while it's completely opposite for scale 3. %This provides good examples of how ASRNNs adapt scales based on different temporal contexts.

\subsection{Word Level Language Modeling}

\begin{table*}[t]
\caption{Perplexities for word level language modeling on WikiText-2 dataset. Italic numbers are reported by original papers.}
\label{tab:Text_pelp}

\begin{center}
\begin{small}
\begin{sc}
\begin{tabular}{lcccccc}
\hline
Methods & $N$ & \specialcell{\# of\\weights} & \specialcell{Min.\\Scale} & \specialcell{Max.\\Scale} & \specialcell{Avg.\\Scale} & PPL \\
\hline\hline
%\abovespace
LSTM & $1024$ & $\approx 10$M & $0$ & $0$ & $0$ & $101.1$ \\
SLSTM & $1024$ & $\approx 10$M & $3$ & $3$ & $3$ & $97.7$ \\
ASLSTM & $1024$ & $\approx 10$M & $0$ & $3$ & $1.51$ & $\mathbf{93.8}$ \\
\hline
GRU & $1024$ & $\approx 7.8$M & $0$ & $0$ & $0$ & $99.7$ \\
SGRU & $1024$ & $\approx 7.8$M & $3$ & $3$ & $3$ & $95.4$ \\
ASGRU & $1024$ & $\approx 7.8$M & $0$ & $3$ & $1.38$ & $\mathbf{92.6}$ \\
\hline\hline
Zoneout + Variational LSTM \cite{wikitext2} & - & - & - & - & - & $100.9$ \\
Pointer Sentinel LSTM \cite{wikitext2} & - & - & - & - & - & $80.8$ \\
Neural cache model \cite{NCM} & $1024$ & - & - & - & - & $\mathbf{68.9}$ \\
\hline

\end{tabular}
\end{sc}
\end{small}
\end{center}
%\vskip -0.2in
\end{table*}

\begin{figure*}[t!]
  \centering
    \includegraphics[width=1.\textwidth]{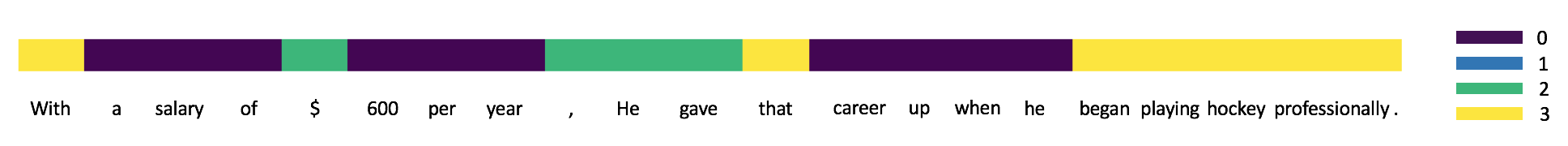}
   \caption{Visualized scale variations for a sampled sentence form WikiText-2 dataset.}
\label{fig:text_scale}
\end{figure*}

Finally, we evaluate ASRNNs for the word level language modeling (WLLM) task on the WikiText-2 \cite{wikitext2} dataset, which contains $2$M training tokens with a vocabulary size of $33$k. We use perplexity as the evaluation metric and the results are summarized in the Table \ref{tab:Text_pelp}, which shows ASRNNs can also outperform their regular counterparts. Besides, Figure \ref{fig:text_scale} further visualizes captured scale variations for a sampled sentence. It indicates scales are usually changed at some special tokens (like semicolon and clause), which comfirms the flexibility of modeling dynamic scale patterns with ASRNNs. What's more, although state-of-the-art models \cite{wikitext2} \cite{NCM} perform better, their techniques are orthogonal to our scaling mechanism so ASRNNs can still benefit from them.

%indicating ASRNNs do adjust scales based on different contents.

\section{Conclusion}\label{sec:conclusion}

We present Adaptively Scaled Recurrent Neural Networks (ASRNNs), a simple yet useful extension that brings dynamical scale modeling abilities to existing RNN structures. At each time step, ASRNNs model the scale patterns by taking causal convolutions between wavelet kernels and input sequences such that the scale can be represented by wavelet scale coefficients. These coefficients are sampled from Gumbel-Softmax (GM) distributions which are parameterized by previous hidden states and current inputs. The differentiable nature of GM allows ASRNNs to learn to adjust scales based on different temporal contexts. Compared with other multiscale RNN models, ASRNNs don't rely on hierarchical architectures and prefixed scale factors, making them simple and easy to train. Evaluations on various sequence modeling tasks indicate ASRNNs can outperform those non-dynamically scaled baselines by adjusting scales according to different temporal information.

\section{Acknowledgment}
This research was partially supported by NSF grant \#1704309.

{\small
\bibliographystyle{aaai}
\bibliography{aaai_2019}
}

\end{document}